\documentclass[runningheads]{llncs}

\usepackage[T1]{fontenc}
\usepackage{microtype}
\usepackage{newtxtext}
\usepackage[varvw]{newtxmath}

\usepackage{amsmath,amsfonts,amssymb}
\usepackage{mathtools}
\usepackage{amstext}

\usepackage{graphicx}
\usepackage{float}

\usepackage{booktabs}
\usepackage{multirow}

\usepackage{algorithm}
\usepackage{algpseudocode}

\usepackage{xcolor}
\usepackage{hyperref}
\usepackage{xspace}
\usepackage{url}
\usepackage{comment}
\newcommand{\egmhql}{Q($\gamma_H$)\xspace} 
\newcommand{\qagg}{Q_{\mathrm{agg}}}

\begin{document}

\title{Q($\gamma_H$): Entropy-Guided Horizon Mixing\\
       in Multi-Discount Q-Learning}

\authorrunning{Arwa Albelaihi and Abdulrahman Altahhan}

\institute{School of Computer Science, University of Leeds,
           Leeds LS2 9JT, UK\\
           \email{\inst{1}Scaaalb@leeds.ac.uk, \inst{2}A.Altahhan@leeds.ac.uk}}

\author{Arwa Albelaihi\inst{1}\orcidID{0000-0001-7135-5894}
\and
Abdulrahman Altahhan\inst{1}\orcidID{0000-0003-1133-7744}}

\authorrunning{A. Albelaihi and A. Altahhan}

\institute{School of Computer Science, University of Leeds,
           Leeds LS2 9JT, UK\\
           \email{scaaalb@leeds.ac.uk, a.altahhan@leeds.ac.uk}}

\maketitle

\begin{abstract}
A fixed discount factor in Q-learning imposes a single planning horizon throughout training, despite the differing demands of early exploration and long-term credit assignment. We propose \egmhql, a multi-horizon Q-learning framework that maintains multiple discount-specific heads and adaptively combines them at control time using an entropy-derived confidence signal. The resulting state-dependent mixing weights shift control from shorter- to longer-horizon heads as policy confidence increases, enabling adaptive horizon mixing without modifying individual Bellman updates. Across classical control and deep reinforcement-learning benchmarks, \egmhql shows its strongest improvements on delayed-credit tasks relative to strong fixed-discount baselines. By decoupling horizon adaptation from the underlying learning updates, \egmhql provides a lightweight framework for adaptive planning-horizon control in reinforcement learning, avoiding the need for manual discount-factor selection while remaining compatible with standard value-learning methods.

\end{abstract}

\section{Introduction}
\label{sec:intro}
Choosing the discount factor $\gamma$ in reinforcement learning
is challenging. A short horizon can stabilise early learning by limiting
bootstrap error propagation, but may prevent solving tasks requiring
long-range credit assignment
\cite{fedus2019hyperbolic,romoff2019separating}.

No single fixed $\gamma$ resolves this tension, and per-task tuning
\cite{henderson2018deep} does not address changes in the required
planning horizon during training. Neuroscience evidence also suggests
that dopaminergic neurons encode reward prediction errors across
multiple discount time constants \cite{masset2025nature}.

Recent methods address this through adaptive or multi-horizon
discounting
\cite{ogawa2026adaptive,ali2023multi,nafi2024horizon,moghimi2026decoupling},
but require modified Bellman targets, global rather than state-dependent
mixing, reward decomposition, or added deep-RL infrastructure.
These approaches do not provide state-dependent control-time mixing
across discount-specific Q-heads while preserving their standard
Q-learning updates.
We propose \egmhql, which maintains $K$ heads with discounts
$\gamma_k \in \Gamma$ and mixes them using state-dependent confidence.
Low confidence favours shorter horizons, while high confidence favours
longer ones. For analysis, we define
$\bar{\gamma}(s)=\sum_{k=1}^{K}w_k(s)\gamma_k$ as a
mixture-weighted discount summary of horizon emphasis, not the discount
of an equivalent Bellman operator. Each head retains its fixed-$\gamma_k$ Q-learning update.

Across tabular, linear, and deep RL settings, \egmhql is strongest on
delayed-credit tasks while remaining competitive across the
other evaluated environments.
Our contributions are: (i) state-dependent horizon mixing that preserves
standard Q-learning updates; (ii) a deep RL extension with
discount-specific heads; and (iii) controlled and ablation analyses of
the entropy-guided mixing mechanism.

\section{Related Work}

\egmhql builds on multi-horizon value learning, adaptive discounting,
and entropy-based uncertainty signals. Unlike prior work, it performs state-dependent
mixing across planning horizons only at control time, preserves standard Bellman updates,
and extends to deep RL through discount-specific Q-heads.

Parallel value prediction across planning horizons is central to the GVF
framework~\cite{sutton2011horde} and discount-indexed value learning
\cite{fedus2019hyperbolic,romoff2019separating}. Related deep-RL methods
use multiple discount-indexed Q-heads~\cite{ali2023multi},
reward-component-specific discounts~\cite{chen2026multidiscounting},
multi-discount PPO advantages~\cite{nafi2024horizon}, or time-weighted
horizon heads in distributional RL~\cite{moghimi2026decoupling}. \egmhql
also uses discount-specific heads, but couples them only at action selection:
it requires no reward decomposition~\cite{chen2026multidiscounting}, uses
per-state rather than global mixing~\cite{nafi2024horizon}, and avoids
distributional RL machinery~\cite{moghimi2026decoupling}.

Prior work has formalised transition- and state-dependent discounting
$\gamma(s,a,s')$ ~\cite{white2017unifying,sharma2021transition}, used discounts to model termination or continuation values
\cite{yoshida2013reinforcement,stachurski2021dynamic}, and adapted
$\gamma$ online using critic loss or entropy-based uncertainty
\cite{ogawa2026adaptive,cho2025entropy}. In contrast, \egmhql does not
modify Bellman targets or learn $\gamma(s)$ directly; it performs
state-dependent control-time mixing while each head keeps a standard
fixed-$\gamma_k$ update. The resulting
$\bar{\gamma}(s)=\sum_k w_k(s)\gamma_k$ is used only as a descriptive
summary of the mixing weights.

\label{sec:related:entropy}

Entropy is commonly used in RL as an objective
\cite{haarnoja2018soft,ziebart2008maximum}, while ensembles use model
diversity for uncertainty estimation and exploration
\cite{osband2016deep,lakshminarayanan2017simple}, and
mixture-of-experts methods use input-dependent gating
\cite{shazeer2017outrageously}. In contrast, \egmhql uses reference-policy
entropy as a lightweight signal for decision ambiguity. Its head diversity is
temporal rather than epistemic, and the gate interpolates across planning
horizons rather than averaging to reduce variance. Table~\ref{tab:related}
summarises the closest comparisons.

A complementary motivation comes from neuroscience: dopaminergic neurons
have been shown to encode reward prediction errors across multiple
discount time constants~\cite{masset2025nature}. This supports the use
of parallel fixed-$\gamma$ estimators, while \egmhql adds an adaptive
aggregation mechanism through entropy-guided control-time mixing.

\begin{table}[t]
\centering
\caption{Comparison of \egmhql with closely related methods.
         State-dep.\ = state-dependent mixing or adaptation.
         Bellman changed = whether Bellman targets are modified.
         Control mix = whether discount-specific heads are mixed
         at action selection.}
\label{tab:related}
\renewcommand{\arraystretch}{1}
\setlength{\tabcolsep}{1pt}
\begin{tabular}{lccccc}
\hline
\textbf{Method} &
\textbf{Multi-} & \textbf{State-} &
\textbf{Gate} &
\textbf{Bellman} & \textbf{Control} \\
&
\textbf{head} & \textbf{dep.} &
\textbf{signal} &
\textbf{changed} & \textbf{mix} \\
\hline
\egmhql (ours)
    & \checkmark & \checkmark & Entropy & No & \checkmark \\
Ogawa et al.\ \cite{ogawa2026adaptive}
    & $\times$ & $\times$ & Critic loss & Yes & $\times$ \\
Moghimi et al.\ \cite{moghimi2026decoupling}
    & \checkmark & $\times$ & None & No & $\times$ \\
Nafi et al.\ \cite{nafi2024horizon}
    & \checkmark & $\times$ & None & No & $\times$ \\
Chen et al.\ \cite{chen2026multidiscounting}
    & \checkmark & $\times$ & None & No & $\times$ \\
\hline
\end{tabular}
\end{table}

\section{Method: Q($\gamma_H$)}
\label{sec:method}

\subsection{Problem Setup}
\label{sec:method:motivation}
We consider a finite Markov decision process
$\mathcal{M}=(\mathcal{S},\mathcal{A},P,r,\gamma)$ with finite action
space $\mathcal{A}$ and discount factor $\gamma\in(0,1)$. Standard
Q-learning~\cite{watkins1992qlearning} fixes $\gamma$ before training,
although the appropriate planning horizon may vary across states and
learning stages. \egmhql instead maintains $K$ parallel Q-learning heads,
each trained with a different fixed discount factor, and mixes their
values at control time using an entropy-derived confidence signal.
\subsection{Multi-Horizon Q-Heads}
\label{sec:method:heads}

Let $\Gamma=\{\gamma_1<\cdots<\gamma_K\}$ be a set of discount factors.
For each $\gamma_k$, \egmhql maintains an independent Q-head
$Q^{(k)}:\mathcal{S}\times\mathcal{A}\to\mathbb{R}$:
\begin{equation}
    Q^{(k)}(s,a)\approx
    \mathbb{E}\!\left[\sum_{t=0}^{\infty}\gamma_k^t r_t
    \mid s_0=s,a_0=a\right].
    \label{eq:qhead}
\end{equation}
Each head is parameterised as
\begin{equation}
    Q^{(k)}(s,a)=W_k(a,:)^\top\phi(s),
    \label{eq:linear}
\end{equation}
with its independent discount factor $\gamma_k$ and Bellman targets, while sharing the same
experience stream and feature representation.


\subsection{Reference Q Construction}
\label{sec:method:reference}

To compute confidence, we construct a reference Q-vector
$q_{\mathrm{ref}}\in\mathbb{R}^{|\mathcal{A}|}$ from the current
multi-head estimates. A simple baseline uses one fixed head:
\begin{equation}
    q_{\mathrm{ref}} = Q^{(k_{\mathrm{ref}})}(s,\cdot),
    \label{eq:fixed_ref}
\end{equation}
which we evaluate in the ablation study.

Our default uses all heads through a max--min aggregation:
\begin{equation}
    q_{\mathrm{ref}}(a)
    =
    \eta \max_k Q^{(k)}(s,a)
    +
    (1-\eta)\min_k Q^{(k)}(s,a),
    \qquad \forall a\in\mathcal{A},
    \label{eq:maxmin}
\end{equation}
where $\eta\in[0,1]$ balances optimistic and conservative estimates. We
use $\eta=0.5$ by default and assess alternatives in the ablation and
sensitivity studies.

\subsection{Entropy-Based Confidence}
\label{sec:method:entropy}
Given $q_{\mathrm{ref}}$, we compute a Boltzmann policy with temperature
$\tau > 0$:
\begin{equation}
    \pi_{\mathrm{ref}}(a\mid s)
    =
    \frac{\exp(q_{\mathrm{ref}}(a)/\tau)}
         {\sum_{a'}\exp(q_{\mathrm{ref}}(a')/\tau)}.
    \label{eq:boltzmann}
\end{equation}
The normalised Shannon entropy of this policy is:
\begin{equation}
    H_n(s)
    =
    -\frac{1}{\log|\mathcal{A}|}
    \sum_{a\in\mathcal{A}}
    \pi_{\mathrm{ref}}(a\mid s)\log \pi_{\mathrm{ref}}(a\mid s)
    \label{eq:entropy}
\end{equation}
Normalisation by $\log|\mathcal{A}|$ ensures $H_n(s) = 1$ when all
actions appear equally valuable and $H_n(s) \to 0$
when one action is clearly preferred. The agent's
\emph{confidence} is then:
\begin{equation}
    c(s)=1-H_n(s).
    \label{eq:confidence}
\end{equation}
Thus, high entropy implies low confidence, while low entropy implies high confidence.

\subsection{Adaptive Horizon Mixing}
\label{sec:method:mixing}

We map confidence $c(s) \in [0,1]$ to a distribution over the $K$ heads
using a soft proximity weighting. Each head $k$ is assigned a fixed
position on the unit interval:
\begin{equation}
    u_k = \frac{k - 1}{K - 1}, \qquad k = 1, \ldots, K,
    \label{eq:headpos}
\end{equation}
so that $u_1 = 0$ corresponds to the shortest-horizon head and $u_K = 1$
to the longest-horizon head. Mixing weights are computed as a softmax
over negative distances from the confidence value:
\begin{equation}
    w_k(s)
    \;=\;
    \frac{\exp\!\bigl(-\beta\,|u_k - c(s)|\bigr)}
         {\sum_{j=1}^{K} \exp\!\bigl(-\beta\,|u_j - c(s)|\bigr)},
    \label{eq:weights}
\end{equation}
where $\beta$ controls selection sharpness. Low confidence gives more
weight to short-horizon heads, while high confidence favours longer
horizons.

The head values are aggregated for action selection using the
confidence-modulated weights:
\begin{equation}
    \qagg(s,a)=\sum_{k=1}^{K}w_k(s)Q^{(k)}(s,a).
    \label{eq:qagg}
\end{equation}
For analysis, we additionally define the mixture-weighted discount
summary
\begin{equation}
    \bar{\gamma}(s)=\sum_{k=1}^{K}w_k(s)\gamma_k.
    \label{eq:gbar}
\end{equation}
This quantity summarises whether the mixing weights emphasise shorter-
or longer-horizon heads; it is not the discount factor of an equivalent
Bellman operator. 
\subsection{Learning Rule}
\label{sec:method:learning}

Actions are selected $\varepsilon$-greedily with respect to $\qagg$.
After observing $(s,a,r,s')$, each head is updated independently:
\begin{equation}
    \delta_k =
    r + \mathbf{1}_{\{s' \notin \mathcal{T}\}}\gamma_k
    \max_{a'} Q^{(k)}(s',a')
    - Q^{(k)}(s,a).
    \label{eq:td}
\end{equation}
\begin{equation}
    W_k(a,:) \leftarrow W_k(a,:) + \alpha\delta_k\phi(s).
    \label{eq:update}
\end{equation}
The mixing weights affect only action selection; each head keeps its own
standard fixed-$\gamma_k$ Q-learning update. Algorithm~\ref{alg:egmhql} summarises the full procedure.

\begin{algorithm}[t]
\caption{\egmhql: Entropy-Guided Horizon Mixing}
\label{alg:egmhql}
\begin{algorithmic}[1]
\Require $\Gamma=\{\gamma_1,\ldots,\gamma_K\}$, $\alpha$,
         $\tau$, $\beta$, $\eta$, $\varepsilon$
\State Initialise $W_k \leftarrow \mathbf{0}$ for all $k$;
       set $u_k \leftarrow (k-1)/(K-1)$
\For{each episode}
    \State Observe initial state $s$
    \While{$s$ not terminal}
        \State Compute $Q^{(k)}(s,\cdot) \leftarrow W_k\phi(s)$
               for all $k$
        \State Build reference:
               $q_{\mathrm{ref}}(a) \leftarrow
               \eta\max_k Q^{(k)}(s,a) +
               (1{-}\eta)\min_k Q^{(k)}(s,a)$
        \State Derive $\pi_{\mathrm{ref}}(\cdot\mid s)$ from
               $q_{\mathrm{ref}}$ via softmax with $\tau$;
               compute entropy $H_n(s)$
        \State Set $c(s) \leftarrow 1 - H_n(s)$; compute
               weights $w_k(s)$ via softmax over $|u_k - c(s)|$
        \State $\qagg(s,\cdot) \leftarrow
               \sum_k w_k(s)\,Q^{(k)}(s,\cdot)$
        \State Select $a$ $\varepsilon$-greedily w.r.t.\
               $\qagg(s,\cdot)$; execute; observe $r$, $s'$
         \State $\delta_k = r + \mathbf{1}_{\{s' \notin \mathcal{T}\}}\,\gamma_k
       \max_{a'} Q^{(k)}(s',a') - Q^{(k)}(s,a)$ \quad for all $k$    
        \State $W_k(a,:) \leftarrow
               W_k(a,:) + \alpha\,\delta_k\,\phi(s)$
               \quad for all $k$
        \State $s \leftarrow s'$
    \EndWhile
    \State Decay $\varepsilon$
\EndFor
\end{algorithmic}
\end{algorithm}

\subsection{Complexity and Practical Considerations}
\label{sec:method:theory}

Each head in \egmhql uses a
standard off-policy Q-learning update under its own Bellman operator,
where sufficient exploration is the key practical requirement
\cite{watkins1992qlearning,tsitsiklis1994asynchronous}. The mixing
weights are bounded, non-negative, and sum to one, so $\qagg$ is a
well-defined convex combination of the heads:
\[
\qagg(s,a)=\sum_k w_k(s)Q^{(k)}(s,a).
\]
Thus, state-dependent mixing affects action selection but not the individual Bellman
targets. 

The overhead of \egmhql is linear in the number of heads $K$.
Maintaining $K$ heads requires $O(K|\mathcal{A}|d)$ memory, and dense
linear evaluation and updates cost $O(K|\mathcal{A}|d)$ per step. The
reference, entropy, and weighting computations add only
$O(K|\mathcal{A}|)$ operations. In tabular and linear settings, no replay
buffer, target network, or neural component is required. In the deep-RL
extension, the same confidence-gated mixing is implemented on top of
standard DQN and DDQN machinery using discount-specific output heads.

\section{Experiments}
\label{sec:experiments}
We evaluate \egmhql in tabular/linear and deep-RL settings.
The tabular/linear suite contains eight benchmarks spanning delayed-credit,
navigation, stochastic, and risk-sensitive tasks. The deep learning setting
evaluates CartPole-v1, Acrobot-v1, and MountainCar-v0 using a shared-trunk
multi-head DDQN architecture with one head per $\gamma \in \Gamma$.
Table~\ref{tab:environments} summarises all environments.

In the tabular and linear settings, we compare \egmhql with Q-learning,
SARSA, Expected SARSA, and Double Q-learning. Each baseline uses an oracle
discount $\gamma^*$ selected post hoc from $\Gamma$ using best performance,
while \egmhql uses all discounts in $\Gamma$ and adapts its horizon mixture
online.

\begin{table}[t]
\centering
\caption{Experimental environments used in evaluations. TC denotes tile coding, IHT denotes index hash table, and FC denotes a fully connected neural network.}
\label{tab:environments}
\footnotesize
\renewcommand{\arraystretch}{1.1}
\setlength{\tabcolsep}{3pt}
\begin{tabular}{llccl}
\hline
\textbf{Environment} & \textbf{Type} &
\textbf{$|\mathcal{A}|$} & \textbf{Repr.} &
\textbf{Key challenge} \\
\hline
\multicolumn{5}{l}{\textit{Tabular / linear}} \\
\hline
MountainCar & Continuous & 3 & TC (IHT 2048) & Delayed credit \\
Acrobot  & Continuous & 3 & TC (IHT 4096) & Delayed credit \\
LunarLander           & Continuous & 4 & TC (IHT 8192) & Dense reward \\
Gridworld             & Navigation & 4 & Tabular        & Deterministic \\
WindyGridworld        & Navigation & 8 & Tabular        & Stochastic wind \\
CliffWalking          & Navigation & 4 & Tabular        & Cliff penalty \\
Stoch.\ CliffWalking  & Navigation & 4 & Tabular        & Slip $p{=}0.10$ \\
FrozenLake-8$\times$8 & Stoch.\ nav. & 4 & One-hot     & Slippery, sparse \\
\hline
\multicolumn{5}{l}{\textit{Deep learning (normalised state)}} \\
\hline
CartPole-v1              & Control & 2 & FC $[128,128,64]$ & Easy baseline \\
Acrobot-v1     & Control & 3 & FC $[128,128,64]$ & Delayed credit \\
MountainCar-v0 & Control & 3 & FC $[128,128,64]$ & Hard exploration \\
\hline
\end{tabular}
\end{table}

\begin{table}[t]
\centering
\caption{Default hyperparameters used in the experiments.}
\label{tab:hyperparams}
\footnotesize
\renewcommand{\arraystretch}{1.1}
\setlength{\tabcolsep}{3pt}
\begin{tabular}{llccl}
\hline
\textbf{Parameter} & \textbf{Symbol} & \textbf{Value} & \textbf{Notes} \\
\hline
\multicolumn{4}{l}{\textit{Shared (tabular/linear \& deep)}} \\
Initial exploration & $\varepsilon_0$      & $0.10$  & $\varepsilon$-greedy \\
Exploration decay   & $\delta_\varepsilon$ & $0.985$ & Per-episode \\
Min.\ exploration   & $\varepsilon_{\min}$ & $0.05$  & \\
\hline
\multicolumn{4}{l}{\textit{Tabular/linear only}} \\
Discount set    & $\Gamma$ & $\{0.70,0.80,0.90,0.97,0.99\}$ & $K=5$ heads \\
Learning rate   & $\alpha$ & $0.025$  & TD step size \\
Episodes        & $T$      & $200$    & \\
Max steps       & \texttt{max\_t} & $5{,}000$ & \\
Temperature     & $\tau$   & $0.25$   & \\
Sharpness       & $\beta$  & $3.0$    & \\
Mixing          & $\eta$   & $0.5$    & \\
\hline
\multicolumn{4}{l}{\textit{Deep learning only}} \\
Discount set    & $\Gamma$ & $\{0.90, 0.95, 0.99\}$ & $K=3$ heads \\
Learning rate   & $\alpha$ & $10^{-4}$ & Adam optimiser \\
Episodes        & $T$      & $300$  &  \\
Replay buffer   & —        & $50{,}000$ & Uniform sampling \\
Batch size      & —        & $64$ & \\
Target sync     & —        & Every $100$ steps & Hard update \\
Entropy temp.   & $\tau$   & $0.1$ & Sharper than tabular \\
Entropy warmup  & —        & $200$ steps & \\
Network         & —        & FC $[128, 128, 64]$ & Shared trunk \\
\hline
\end{tabular}
\end{table}

In the deep-RL setting, we compare Deep~Q($\gamma$H) against five
baselines: DDQN~\cite{hasselt2016deep} and
DQN~\cite{mnih2015human} (single-head, fixed $\gamma$);
Hyperbolic DQN~\cite{fedus2019hyperbolic} ($K$ heads with
hyperbolic spacing $\gamma_k{=}1/(1{+}kh)$, uniform mixing, no
confidence gate); Ogawa-Adapt~\cite{ogawa2026adaptive}
(single $\gamma$ adjusted by a loss-based PI controller); and
Scheduled-$\gamma$ ($\gamma$ grown linearly from $0.80$ to $0.99$,
no learned signal).

Table~\ref{tab:hyperparams} lists all hyperparameters. Shared parameters
are fixed across environments and both settings. Unless otherwise stated,
experiments use $N{=}10$ seeds. For MountainCar, the default \egmhql
configuration uses the same 10 canonical runs wherever the identical
setting appears; alternative configurations use separate 10-seed runs,
with 20-seed results reported where stated. We report EarlyMean (first
50 episodes), TailMean (last 50 episodes), and AUC (full run). Episode
length is the performance metric (lower is better) except for
LunarLander (return, higher is better).

\subsection{Main Results}
\label{sec:experiments:main}
Table~\ref{tab:all_results} shows that \egmhql is strongest on
delayed-credit tasks. It achieves the best mean AUC on MountainCar and
Acrobot, with reductions of 6.0\% and 2.1\%, respectively, relative
to the next-best fixed-$\gamma$ baselines. Across the remaining environments,
\egmhql remains competitive and often improves early learning, although
fixed-$\gamma$ or on-policy baselines can perform better when one
horizon is sufficient or risk-sensitive behaviour is required.

Table~\ref{tab:deep_all} extends the comparison to deep RL.
Deep~Q($\gamma$H)-DDQN achieves the best AUC on all three environments.
On Acrobot-v1 and MountainCar-v0, it outperforms DDQN by 8.5\% and
7.0\%, respectively. The largest margins are against adaptive baselines:
Scheduled-$\gamma$ is outperformed by 48.0\% on Acrobot
($p{<}0.0001$) and 7.5\% on MountainCar ($p{=}0.0007$), while
Ogawa-Adapt is outperformed by 21.4\% ($p{=}0.0029$) and 7.6\%
($p{=}0.0008$), respectively. Within these classic-control tasks,
these results support the benefit of confidence-guided horizon mixing.

\subsection{Mechanism Analysis}
\label{sec:experiments:mechanism}

\begin{figure}[t]
    \centering
    \includegraphics[width=\linewidth]
        {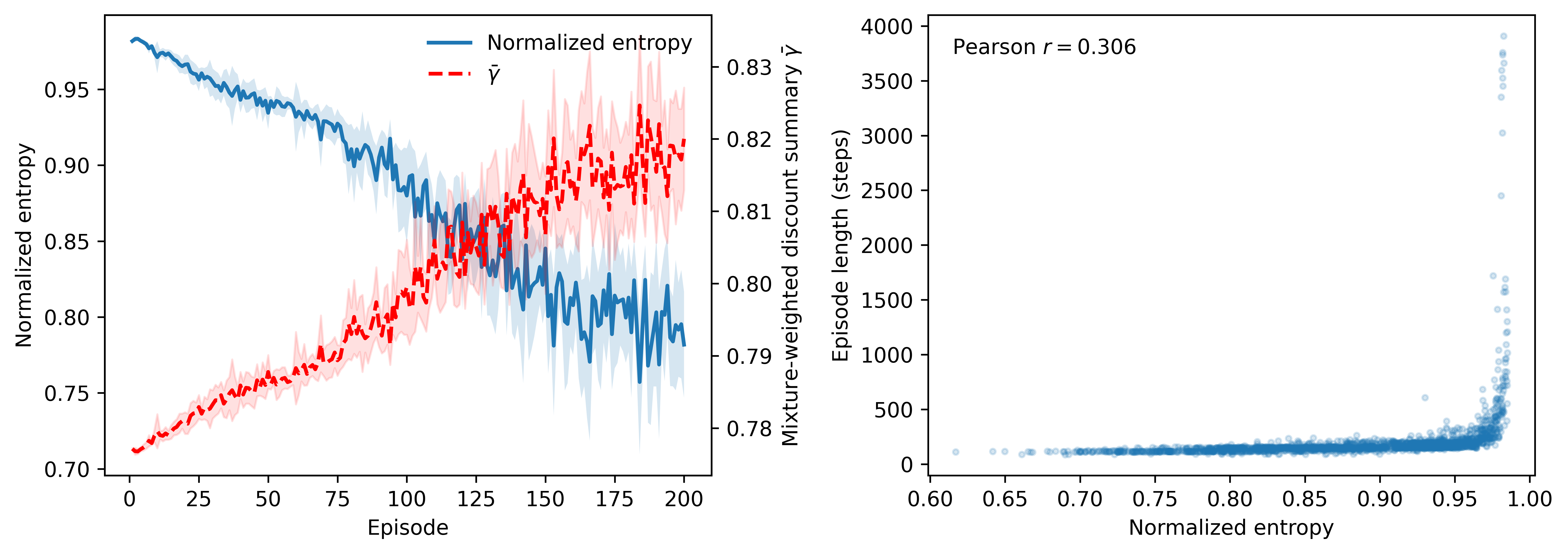}
    \caption{Mechanism analysis on MountainCar using the same 10 canonical
training seeds. Left: normalised entropy decreases while
$\bar{\gamma}$ increases; shaded bands denote 95\% confidence intervals
across seeds. Right: normalised entropy versus episode length across all
seed--episode observations ($N=2{,}000$, Pearson $r=0.306$).}
    \label{fig:mechanism_evolution}
\end{figure}

Figure~\ref{fig:mechanism_evolution} examines the internal mixing
signals on MountainCar. As training progresses, normalised entropy
decreases from $0.960$ to $0.800$, while $\bar{\gamma}$ increases
from $0.782$ to $0.816$, indicating a shift in mixing weight toward
longer-horizon heads. The entropy--episode length relationship shows a
positive association ($r=0.306$), with higher entropy associated with
longer episodes and poorer performance, consistent with its role as a
lightweight confidence signal for horizon mixing.

To examine state dependence directly, we evaluated the learned gate over
the MountainCar position--velocity state space using the same 10 canonical
training seeds. In the 10-seed mean state-space evaluation, $H_n$ ranged
from $0.042$ to $1.000$ and $\bar{\gamma}$ from $0.773$ to $0.951$.
The weights of the $\gamma=0.70$ and $\gamma=0.99$ heads ranged from
$0.031$--$0.540$ and $0.027$--$0.479$, respectively, showing that the
learned horizon mixture varies substantially across states.

\begin{table}[H]
\centering
\footnotesize
\caption{\egmhql vs.\ oracle-tuned fixed-$\gamma$ baselines.
Values are means over 10 seeds. Steps $\downarrow$ except
LunarLander return $\uparrow$.}
\label{tab:all_results}
\renewcommand{\arraystretch}{1.1}
\setlength{\tabcolsep}{2pt}
{%
\begin{tabular}{llcccc}
\hline
\textbf{Environment} & \textbf{Method} & $\gamma^*$ &
\textbf{Early} & \textbf{Tail} & \textbf{AUC} \\
\hline
\multicolumn{6}{l}{\textit{Delayed-credit environments}} \\
\hline
\multirow{5}{*}{MountainCar}
    & \egmhql       & —    & \textbf{366.36} & 136.45 & \textbf{203.58} \\
    & Q-Learning    & 0.97 & 402.42 & \textbf{136.16} & 217.12 \\
    & SARSA         & 0.97 & 388.81 & 143.95 & 216.51 \\
    & Exp.\ SARSA   & 0.97 & 401.04 & 139.79 & 218.13 \\
    & Double Q      & 0.97 & 580.43 & 146.54 & 280.95 \\
\hline
\multirow{5}{*}{Acrobot}
    & \egmhql       & —    & 602.97 & \textbf{127.88} & \textbf{192.78} \\
    & Q-Learning    & 0.97 & 535.94 & 144.53 & 200.92 \\
    & SARSA         & 0.97 & 543.22 & 144.99 & 200.32 \\
    & Exp.\ SARSA   & 0.99 & \textbf{521.68} & 143.68 & 196.96 \\
    & Double Q      & 0.97 & 636.53 & 163.37 & 228.16 \\
\hline
\multicolumn{6}{l}{\textit{General environments}} \\
\hline
\multirow{5}{*}{Gridworld}
    & \egmhql       & —    & \textbf{65.88} & 19.60 & 37.60 \\
    & Q-Learning    & 0.99 & 66.14 & 19.74 & \textbf{36.96} \\
    & SARSA         & 0.97 & 68.68 & \textbf{19.32} & 37.16 \\
    & Exp.\ SARSA   & 0.97 & 66.16 & 19.96 & 36.98 \\
    & Double Q      & 0.99 & 77.70 & 31.74 & 48.90 \\
\hline
\multirow{5}{*}{WindyGridworld}
    & \egmhql       & —    & \textbf{319.23} & 93.78 & 162.93 \\
    & Q-Learning    & 0.70 & 369.04 & \textbf{68.37} & \textbf{160.50} \\
    & SARSA         & 0.80 & 347.49 & 73.59 & 162.65 \\
    & Exp.\ SARSA   & 0.80 & 354.09 & 72.73 & 163.10 \\
    & Double Q      & 0.70 & 594.94 & 108.61 & 268.34 \\
\hline
\multirow{5}{*}{FrozenLake}
    & \egmhql       & —    & \textbf{197.40} & 22.34 & 76.70 \\
    & Q-Learning    & 0.97 & 197.72 & 22.16 & 76.60 \\
    & SARSA         & 0.99 & 198.22 & \textbf{21.02} & 77.13 \\
    & Exp.\ SARSA   & 0.80 & 197.97 & 21.36 & \textbf{72.80} \\
    & Double Q      & 0.80 & 198.43 & 25.90 & 82.24 \\
\hline
\multirow{5}{*}{LunarLander $\uparrow$}
    & \egmhql       & —    & \textbf{$-$162.43} & $-$83.53 & $-$122.56 \\
    & Q-Learning    & 0.99 & $-$164.36 & \textbf{$-$74.74} & \textbf{$-$116.09} \\
    & SARSA         & 0.99 & $-$168.46 & $-$102.92 & $-$137.53 \\
    & Exp.\ SARSA   & 0.99 & $-$166.03 & $-$105.40 & $-$138.27 \\
    & Double Q      & 0.99 & $-$171.01 & $-$93.53 & $-$130.96 \\
\hline
\multicolumn{6}{l}{\textit{Risk-sensitive environments}} \\
\hline
\multirow{5}{*}{CliffWalking}
    & \egmhql       & —    & 128.06 & \textbf{24.29} & 58.29 \\
    & Q-Learning    & 0.99 & \textbf{118.57} & 24.31 & \textbf{55.17} \\
    & SARSA         & 0.97 & 120.60 & 25.05 & 57.03 \\
    & Exp.\ SARSA   & 0.70 & 131.43 & 24.30 & 59.42 \\
    & Double Q      & 0.99 & 159.47 & 48.80 & 88.31 \\
\hline
\multirow{5}{*}{Stoch.\ Cliff}
    & \egmhql       & —    & 130.81 & 29.04 & 61.85 \\
    & Q-Learning    & 0.70 & 148.85 & 27.82 & 68.94 \\
    & SARSA         & 0.99 & \textbf{122.11} & \textbf{27.64} & \textbf{58.74} \\
    & Exp.\ SARSA   & 0.80 & 130.94 & 27.34 & 61.85 \\
    & Double Q      & 0.97 & 145.96 & 39.64 & 79.14 \\
\hline
\end{tabular}}
\end{table}

\begin{table}[t]
\centering
\caption{Deep Q($\gamma$H)-DDQN vs.\ all baselines
(10 seeds, 300 episodes). AUC $\uparrow$ higher is better.
Gain is computed as
$100 \times (\mathrm{AUC}_{Q(\gamma H)}-\mathrm{AUC}_{\mathrm{base}})/
|\mathrm{AUC}_{\mathrm{base}}|$;
positive values indicate that Q($\gamma$H)-DDQN performs better.
$^{*}p{<}0.05$,\ $^{**}p{<}0.01$,\ $^{***}p{<}0.001$
(paired $t$-test, 10 seeds).
\textbf{Bold} = best per column.}
\label{tab:deep_all}
\renewcommand{\arraystretch}{1.2}
\setlength{\tabcolsep}{4pt}
\scriptsize
\begin{tabular}{lrrrrrr}
\hline
\multirow{2}{*}{\textbf{Method}} &
\multicolumn{2}{c}{\textbf{CartPole-v1} $\uparrow$} &
\multicolumn{2}{c}{\textbf{Acrobot-v1} $\uparrow$} &
\multicolumn{2}{c}{\textbf{MountainCar-v0} $\uparrow$} \\
\cmidrule(lr){2-3}\cmidrule(lr){4-5}\cmidrule(lr){6-7}
& AUC & Gain & AUC & Gain & AUC & Gain \\
\hline
Q($\gamma$H)-DDQN  & \textbf{40,476} & —
                   & \textbf{$-$34,660} & —
                   & \textbf{$-$55,185} & — \\
Q($\gamma$H)-DQN   & 38,804 & $+4.3\%$
                   & $-$40,242 & $+13.9\%$
                   & $-$58,312 & $+5.4\%^{*}$ \\
DDQN               & 38,084 & $+6.3\%$
                   & $-$37,900 & $+8.5\%$
                   & $-$59,314 & $+7.0\%^{**}$ \\
DQN                & 26,851 & $+50.7\%$
                   & $-$35,634 & $+2.7\%$
                   & $-$59,112 & $+6.6\%^{**}$ \\
Hyperbolic~\cite{fedus2019hyperbolic}
                   & 34,616 & $+16.9\%$
                   & $-$38,169 & $+9.2\%^{**}$
                   & $-$58,025 & $+4.9\%^{*}$ \\
Ogawa-Adapt~\cite{ogawa2026adaptive}
                   & 36,999 & $+9.4\%$
                   & $-$44,102 & $+21.4\%^{**}$
                   & $-$59,702 & $+7.6\%^{**}$ \\
Scheduled-$\gamma$ & 40,226 & $+0.6\%$
                   & $-$66,684 & $+48.0\%^{***}$
                   & $-$59,638 & $+7.5\%^{**}$ \\
\hline
\end{tabular}
\end{table}

\begin{figure}[t]
  \centering
  \includegraphics[width=0.7\linewidth]{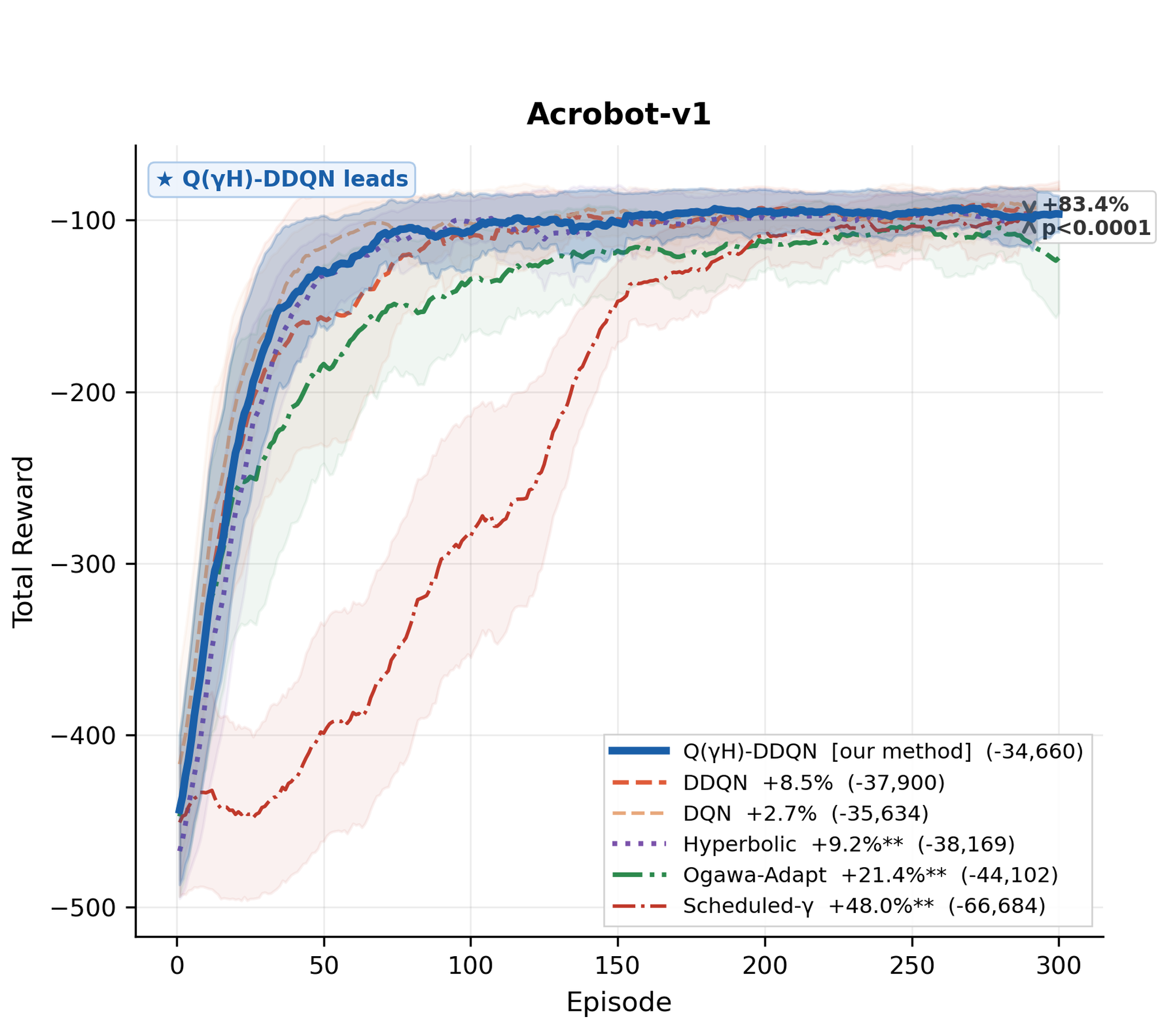}
  \caption{Learning curves on Acrobot-v1 (10 seeds, 300 episodes,
  shaded $= 95\%$ CI, reward $\uparrow$).
  Q($\gamma$H)-DDQN (solid blue) converges faster and achieves
  the best cumulative reward ($-34{,}660$).
  It outperforms Scheduled-$\gamma$ by $48.0\%$ ($p{<}0.001$),
  Ogawa-Adapt by $21.4\%$ ($p{<}0.01$), and DDQN by $8.5\%$.
  The annotation marks the gain over Scheduled-$\gamma$ relative
  to the episode-wise peak gap.
  Since Acrobot assigns $-1$ per timestep, values closer to zero
  indicate faster task completion.}
  \label{fig:deep_all}
\end{figure}

\subsection{Head Disagreement as Uncertainty Proxy}
\label{sec:experiments:mechanism:disagreement}

We also measure head disagreement as an independent value-side diagnostic:
\begin{equation}
    d(s,a)=\max_k Q^{(k)}(s,a)-\min_k Q^{(k)}(s,a).
    \label{eq:disagreement}
\end{equation}
Table~\ref{tab:mechanism_combined} shows that disagreement increases
during learning while entropy decreases and $\bar{\gamma}$ increases.
The negative disagreement--entropy correlations ($r=-0.883$ on
MountainCar and $r=-0.969$ on Acrobot) are consistent with increasingly decisive
reference policies as horizon-specific value estimates separate.
Positive disagreement--$\bar{\gamma}$ correlations indicate a concurrent
shift in mixing weight toward longer-horizon heads.

\begin{table}[t]
\centering
\footnotesize
\caption{Signal dynamics and correlations on delayed-credit tasks
(10 seeds). Pearson $r$ is computed episode-wise over 200 episodes.}
\label{tab:mechanism_combined}
\renewcommand{\arraystretch}{1.2}
\setlength{\tabcolsep}{4pt}
\begin{tabular}{lcc}
\hline
& \textbf{MountainCar} & \textbf{Acrobot} \\
\hline
\multicolumn{3}{l}{\textit{Signal dynamics (first 50 $\to$ last 50 episodes)}} \\
\hline
Mean head disagreement
    & $15.53 \to 27.77$ & $1.03 \to 8.15$ \\
Normalised entropy $H_n$
    & $0.960 \to 0.800$ & $0.969 \to 0.878$ \\
Discount summary $\bar{\gamma}$
    & $0.782 \to 0.816$ & $0.780 \to 0.800$ \\
\hline
\multicolumn{3}{l}{\textit{Pearson $r$ (episode-wise, 200 episodes)}} \\
\hline
Disagreement $\to H_n$
    & $-0.883$ & $-0.969$ \\
Disagreement $\to$ steps
    & $-0.608$ & $-0.318$ \\
Disagreement $\to \bar{\gamma}$
    & $+0.888$ & $+0.970$ \\
\hline
\end{tabular}
\end{table}

\subsection{Ablation Study}
\label{sec:experiments:ablation}

We analyse the contribution of the mixing gate, reference construction,
gating signal, and number of heads.

To isolate the effect of entropy-guided mixing from the general benefit
of maintaining multiple heads, we compare \egmhql with a parameter-matched
uniform multi-head baseline using the same five discounts, representation,
training procedure, and seeds, but fixed weights $w_k=1/K$. \egmhql
improves EarlyMean from $397.63$ to $366.36$, TailMean from $141.83$
to $136.45$, and AUC from $216.50$ to $203.58$. Paired $t$-tests show
significant differences for EarlyMean ($p<0.001$), TailMean
($p=0.007$), and AUC ($p<0.001$). These results show that entropy-guided
mixing improves both learning efficiency and final performance beyond
the general benefit of maintaining multiple discount-specific heads.

We compare fixed-head and aggregated reference constructions on
MountainCar. As shown in Table~\ref{tab:ref_construction}, MaxMin linear
achieves the best EarlyMean, TailMean, and AUC. Mean Q and Max Q also
outperform the best fixed-head variant in AUC, whereas the pessimistic
Min Q construction does not. MaxMin linear improves over the best
fixed-head variant by 2.3\% AUC.

\begin{table}[t]
\centering
\footnotesize
\caption{Reference construction comparison on MountainCar
         (10 seeds, steps $\downarrow$). Final $\bar{\gamma}$ and
         $H_n$ are means over the last 50 episodes.}
\label{tab:ref_construction}
\renewcommand{\arraystretch}{1.2}
\setlength{\tabcolsep}{4pt}
\begin{tabular}{llccccc}
\hline
\textbf{Method} & \textbf{Type} &
\textbf{Early $\downarrow$} & \textbf{Tail $\downarrow$} &
\textbf{AUC $\downarrow$} &
\textbf{$\bar{\gamma}$} & \textbf{$H_n$} \\
\hline
MaxMin linear     & Aggregated
    &  \textbf{366.36} & \textbf{136.45} & \textbf{203.58}
    & 0.816 & 0.800 \\
Mean Q            & Aggregated
    & 373.1 & 137.7 & 206.4 & 0.802 & 0.863 \\
Max Q optimistic  & Aggregated
    & 370.7 & 138.9 & 206.8 & 0.789 & 0.926 \\
Fixed middle      & Fixed head
    & 379.1 & 139.9 & 208.3 & 0.798 & 0.882 \\
Fixed short       & Fixed head
    & 375.4 & 139.2 & 208.5 & 0.789 & 0.924 \\
Min Q pessimistic & Aggregated
    & 379.1 & 140.7 & 209.0 & 0.842 & 0.670 \\
Fixed long        & Fixed head
    & 383.7 & 140.5 & 211.6 & 0.844 & 0.658 \\
\hline
\end{tabular}
\end{table}

We compare three confidence signals on MountainCar using the MaxMin
linear reference: entropy $c(s) = 1 - H_n(s)$,
variance $c(s) = 1/(1+v(s))$, and disagreement
$c(s) = 1/(1+d(s))$, where $v(s)$ and $d(s)$ are the
mean Q-value variance and range across heads, respectively.

\begin{table}[t]
\centering
\footnotesize
\caption{Gating signal ablation on MountainCar (10 seeds,
         MaxMin linear reference, steps $\downarrow$).
         Best value per column in \textbf{bold}.}
\label{tab:signal_ablation}
\renewcommand{\arraystretch}{1.2}
\setlength{\tabcolsep}{4pt}
\begin{tabular}{lccccc}
\hline
\textbf{Signal} &
\textbf{Early $\downarrow$} & \textbf{Tail $\downarrow$} &
\textbf{AUC $\downarrow$} &
\textbf{$\bar{\gamma}$ first 50} & \textbf{$\bar{\gamma}$ last 50} \\
\hline
Entropy
    &  \textbf{366.36} & \textbf{136.45} & \textbf{203.58}
    & 0.782 & 0.816 \\
Variance
    & 372.79 & 138.70 & 205.25
    & 0.795 & 0.784 \\
Disagreement
    & 375.35 & 138.47 & 206.41
    & 0.796 & 0.787 \\
\hline
\end{tabular}
\end{table}

Table~\ref{tab:signal_ablation} shows similar performance across the
three signals, with AUC spanning $203.58$--$206.41$, suggesting that the
mechanism is not tied to one gating signal. Entropy is retained as the default as a bounded measure of action
ambiguity; its scale sensitivity is examined in the sensitivity analysis.

Table~\ref{tab:heads_ablation} compares the default $K=5$ setting with a
smaller $K=3$ variant on MountainCar. In the 10-seed comparison, $K=5$
achieves slightly lower EarlyMean, TailMean, and AUC than $K=3$, although
the differences in EarlyMean and AUC are small. In the separate 20-seed
analysis, $K=3$ performs marginally better. Overall, the small differences
between the two configurations suggest limited sensitivity to the number
of heads in this setting. Both configurations increase $\bar{\gamma}$
over training, indicating a shift in mixing weight toward longer-horizon
heads. We retain $K=5$ as the default to provide finer horizon granularity.

\begin{table}[t]
\centering
\footnotesize
\caption{Head count ablation on MountainCar (steps $\downarrow$),
         10 and 20 seeds. Best value per column in \textbf{bold}.}
\label{tab:heads_ablation}
\renewcommand{\arraystretch}{1.2}
\setlength{\tabcolsep}{4pt}
\begin{tabular}{clcccc}
\hline
\textbf{Seeds} & \textbf{Config} &
\textbf{Early $\downarrow$} & \textbf{Tail $\downarrow$} &
\textbf{AUC $\downarrow$} & \textbf{$\bar{\gamma}$ first/last 50} \\
\hline
\multirow{2}{*}{10}
    & K=3 & 367.20 & 139.39 & 204.16
      & 0.755 / 0.790 \\
    & K=5 & \textbf{366.36} & \textbf{136.45} & \textbf{203.58} & 0.782 / 0.816\\
\hline
\multirow{2}{*}{20}
    & K=3 & \textbf{367.36} & \textbf{139.94} & \textbf{204.62}
      & 0.755 / 0.790 \\
    & K=5 & 370.36 & 140.60 & 205.60 & 0.780 / 0.811 \\
\hline
\end{tabular}
\end{table}

\subsection{Sensitivity Analysis}
\label{sec:experiments:sensitivity}

Table~\ref{tab:sensitivity_summary} summarises sensitivity on
MountainCar. \egmhql is most sensitive to the sharpness parameter
$\beta$: too small a value spreads weight too diffusely and slows early
learning. It is less sensitive to $\tau$ and $\eta$, with robust
performance for $\tau\in[0.25,0.50]$ and $\eta\in[0.5,0.9]$. We therefore
use $\beta=3.0$, $\tau=0.25$, and $\eta=0.5$ as defaults.

\begin{table}[t]
\centering
\footnotesize
\caption{Sensitivity analysis on MountainCar (10 seeds,
         steps $\downarrow$). Robust ranges are determined from the
         corresponding AUC results. Best value per block in \textbf{bold}.}
\label{tab:sensitivity_summary}
\renewcommand{\arraystretch}{1.2}
\setlength{\tabcolsep}{4pt}
\begin{tabular}{lcccccc}
\hline
\textbf{Param} & \textbf{Value} &
\textbf{Early $\downarrow$} & \textbf{Tail $\downarrow$} &
\textbf{Default} & \textbf{Robust range} &
\textbf{Sensitivity} \\
\hline
\multirow{3}{*}{$\beta$}
    & \textbf{3.0} & \textbf{366.36} & \textbf{136.45}
      & \multirow{3}{*}{3.0} & \multirow{3}{*}{$[1.5,\ 3.0]$}
      & \multirow{3}{*}{Moderate} \\
    & 1.5 & 388.69 & 140.0 & & & \\
    & 0.5 & 403.03 & 141.9 & & & \\
\hline
\multirow{3}{*}{$\tau$}
    & \textbf{0.25} & \textbf{366.36} & \textbf{136.45}
      & \multirow{3}{*}{0.25} & \multirow{3}{*}{$[0.25,\ 0.50]$}
      & \multirow{3}{*}{Low} \\
    & 0.50 & 368.12 & 137.8 & & & \\
    & 0.10 & 369.92 & 140.1 & & & \\
\hline
\multirow{5}{*}{$\eta$}
    & \textbf{0.5} & \textbf{366.36} & \textbf{136.45}
      & \multirow{5}{*}{0.5} & \multirow{5}{*}{$[0.5,\ 0.9]$}
      & \multirow{5}{*}{Low--moderate} \\
    & 0.7 & 368.84 & 138.33 & & & \\
    & 0.9 & 373.02 & 137.21 & & & \\
    & 0.3 & 374.86 & 140.68 & & & \\
    & 0.1 & 377.00 & 139.70 & & & \\
\hline
\end{tabular}
\end{table}

\section{Discussion and Limitations}
\label{sec:discussion}

The results show a task-dependent pattern across tabular, linear, and
deep settings: \egmhql is most useful when the planning horizon is
difficult to fix in advance.

The strongest gains occur in MountainCar and Acrobot, where short
horizons support stable early learning but longer horizons are needed for
delayed credit assignment. \egmhql addresses this trade-off by shifting
control from shorter- to longer-horizon heads as confidence increases,
producing a gradual shift in horizon emphasis within a single run.

When a single fixed discount is already effective, as in CartPole and
LunarLander, horizon mixing provides limited benefit. In the cliff
environments, SARSA remains competitive because its on-policy updates
encourage safer behaviour near hazardous states; horizon mixing does not
remove this on-policy advantage.

The deep Q($\gamma$H) results demonstrate that the same control-time
mixing mechanism can be applied with neural function approximation, with
the largest gains on MountainCar-v0 and Acrobot-v1 and little improvement
on CartPole-v1. However, the current deep-RL evaluation is limited to
three classic discrete-action control tasks and a shared-trunk multi-head
architecture, so these results establish applicability rather than broad
deep-RL generality. Further work should evaluate larger benchmark suites,
separate-head architectures, continuous-action actor--critic variants,
and the theoretical properties of the coupled multi-head system.

The tabular and linear experiments isolate the horizon-mixing mechanism
without replay buffers, target networks, or neural optimisation effects.
The comparison is conservative because fixed-$\gamma$ baselines use
oracle-selected $\gamma^*$, whereas \egmhql adapts its horizon mixture
online. In the deep setting, DDQN uses $\gamma=0.99$, the strongest fixed discount considered.

The mechanism analyses make horizon mixing observable through the joint
evolution of confidence, $\bar{\gamma}$, and head disagreement, while the
state-wise analysis confirms that the learned mixture also varies across
the MountainCar state space. Here, $\bar{\gamma}$ is used only as a
descriptive summary of the relative weight assigned to shorter- and
longer-horizon heads.

This pattern is also consistent with neuroscience evidence that reward
prediction errors are represented across multiple discount timescales
\cite{masset2025nature}, although we make no claim of mechanistic
correspondence.

A limitation of the current entropy gate is its sensitivity to reward and
Q-value scale under a fixed Softmax temperature. The reward-scaling
experiment shows that this sensitivity can materially affect performance
at large scale changes. Scale-adaptive temperatures or normalised
confidence signals are therefore important directions for future work.

\section{Conclusion}
\label{sec:conclusion}

We introduced \egmhql, a multi-horizon Q-learning framework that avoids
choosing a single fixed discount factor by maintaining discount-specific
Q-heads and mixing them at control time using a state-dependent
confidence signal. The resulting mixing weights adapt horizon emphasis
during training while each head retains a standard fixed-$\gamma_k$
Q-learning update. For analysis, $\bar{\gamma}(s)=\sum_k
w_k(s)\gamma_k$ provides a descriptive summary of this horizon mixture.

Across eight tabular and linear benchmarks, \egmhql is most effective on
delayed-credit tasks, achieving the best mean  AUC on MountainCar and Acrobot
against oracle-tuned fixed-discount baselines. Mechanism and ablation
analyses show state-dependent horizon mixing and improved learning
efficiency from the entropy-guided gate, while sensitivity analysis
identifies reward/Q-value scale as a limitation of the current
fixed-temperature confidence signal.

The deep-RL extension, Deep Q($\gamma$H), shows the same qualitative
pattern across the three evaluated classic-control tasks, with the
largest gains on MountainCar-v0 and Acrobot-v1 and little improvement on
CartPole-v1. Overall, the results suggest that adaptive horizon mixing is a lightweight and interpretable alternative to manual selection of a
single discount factor, especially when delayed credit assignment makes horizon
choice difficult. Future work will examine larger benchmarks,
continuous-action settings, alternative multi-head architectures, and
scale-robust confidence mechanisms.

\begin{credits}

\subsubsection{\discintname}
The authors have no competing interests to declare.
\end{credits}

\bibliographystyle{splncs04}
\bibliography{main}

@article{hasselt2016deep,
  title={Deep Reinforcement Learning with Double Q-learning},
  author={Hasselt, Hado van and Guez, Arthur and Silver, David},
  journal={Proceedings of the AAAI Conference on Artificial Intelligence},
  year={2016},
  volume={30},
  number={1},
  pages={2094--2100}
}

@article{mnih2015human,
  title={Human-level control through deep reinforcement learning},
  author={Mnih, Volodymyr and Kavukcuoglu, Koray and Silver, David and Rusu, Andrei A and Veness, Joel and Bellemare, Marc G and Graves, Alex and Riedmiller, Martin and Fidjeland, Andreas K and Ostrovski, Georg and others},
  journal={nature},
  volume={518},
  number={7540},
  pages={529--533},
  year={2015},
  publisher={Nature Publishing Group UK London}
}

@article{masset2025nature,
  title={Multi-timescale reinforcement learning in the brain},
  author={Masset, Paul and Tano, Pablo and Kim, HyungGoo R. and Malik, Athar N. and Pouget, Alexandre and Uchida, Naoshige},
  journal={Nature},
  year={2025},
  publisher={Nature Publishing Group},
  doi={10.1038/s41586-025-08929-9}
}

@article{fedus2019hyperbolic,
  title={Hyperbolic discounting and learning over multiple horizons},
  author={Fedus, William and Gelada, Carles and Bengio, Yoshua and Bellemare, Marc G. and Larochelle, Hugo},
  journal={arXiv preprint arXiv:1902.06865},
  year={2019}
}

@article{haarnoja2018soft,
  title={Soft actor-critic algorithms and applications},
  author={Haarnoja, Tuomas and Zhou, Aurick and Hartikainen, Kristian and Tucker, George and Ha, Sehoon and Tan, Jie and Kumar, Vikash and Zhu, Henry and Gupta, Abhishek and Abbeel, Pieter and others},
  journal={arXiv preprint arXiv:1812.05905},
  year={2018}
}

@inproceedings{ali2023multi,
  title={Multi-horizon learning in procedurally-generated environments for off-policy reinforcement learning (student abstract)},
  author={Ali, Raja Farrukh and Duong, Kevin and Nafi, Nasik Muhammad and Hsu, William},
  booktitle={Proceedings of the AAAI Conference on Artificial Intelligence},
  volume={37},
  number={13},
  pages={16150--16151},
  year={2023}
}

@inproceedings{white2017unifying,
  title={Unifying task specification in reinforcement learning},
  author={White, Martha},
  booktitle={International Conference on Machine Learning},
  pages={3742--3750},
  year={2017},
  organization={PMLR}
}

@Article{sharma2021transition,
AUTHOR = {Sharma, Abhinav and Gupta, Ruchir and Lakshmanan, K. and Gupta, Atul},
TITLE = {Transition Based Discount Factor for Model Free Algorithms in Reinforcement Learning},
JOURNAL = {Symmetry},
VOLUME = {13},
YEAR = {2021},
NUMBER = {7},
ARTICLE-NUMBER = {1197},
URL = {https://www.mdpi.com/2073-8994/13/7/1197},
ISSN = {2073-8994},
DOI = {10.3390/sym13071197}
}

@article{cho2025entropy,
  title={Entropy-Guided Distributional Reinforcement Learning with Controlling Uncertainty in Robotic Tasks},
  author={Cho, Hyunjin and Kim, Hyunseok},
  journal={Applied Sciences},
  volume={15},
  number={5},
  pages={2773},
  year={2025},
  publisher={MDPI}
}

@inproceedings{sutton2011horde,
  title={Horde: A scalable real-time architecture for learning knowledge from unsupervised sensorimotor interaction},
  author={Sutton, Richard S and Modayil, Joseph and Delp, Michael and Degris, Thomas and Pilarski, Patrick M and White, Adam and Precup, Doina},
  booktitle={The 10th International Conference on Autonomous Agents and Multiagent Systems-Volume 2},
  pages={761--768},
  year={2011}
}

@inproceedings{romoff2019separating,
  title={Separating value functions across time-scales},
  author={Romoff, Joshua and Henderson, Peter and Touati, Ahmed and Brunskill, Emma and Pineau, Joelle and Ollivier, Yann},
  booktitle={International Conference on Machine Learning},
  pages={5468--5477},
  year={2019},
  organization={PMLR}
}

@article{stachurski2021dynamic,
  title={Dynamic programming with state-dependent discounting},
  author={Stachurski, John and Zhang, Junnan},
  journal={Journal of Economic Theory},
  volume={192},
  pages={105190},
  year={2021},
  publisher={Elsevier}
}

@article{osband2016deep,
  title={Deep exploration via bootstrapped DQN},
  author={Osband, Ian and Blundell, Charles and Pritzel, Alexander and Van Roy, Benjamin},
  journal={Advances in neural information processing systems},
  volume={29},
  year={2016}
}

@article{shazeer2017outrageously,
  title={Outrageously large neural networks: The sparsely-gated mixture-of-experts layer},
  author={Shazeer, Noam and Mirhoseini, Azalia and Maziarz, Krzysztof and Davis, Andy and Le, Quoc and Hinton, Geoffrey and Dean, Jeff},
  journal={arXiv preprint arXiv:1701.06538},
  year={2017}
}

@inproceedings{henderson2018deep,
  title={Deep reinforcement learning that matters},
  author={Henderson, Peter and Islam, Riashat and Bachman, Philip and Pineau, Joelle and Precup, Doina and Meger, David},
  booktitle={Proceedings of the AAAI conference on artificial intelligence},
  volume={32},
  number={1},
  year={2018}
}

@article{watkins1992qlearning,
  author  = {Watkins, Christopher J. C. H. and Dayan, Peter},
  title   = {Q-learning},
  journal = {Machine Learning},
  volume  = {8},
  number  = {3--4},
  pages   = {279--292},
  year    = {1992}
}

@inproceedings{nafi2024horizon,
  author    = {Nafi, Nasik Muhammad and Ali, Raja Farrukh and Hsu, William
               and Duong, Kevin and Vick, Mason},
  title     = {Policy Optimization using Horizon Regularized Advantage to
               Improve Generalization in Reinforcement Learning},
  booktitle = {Proceedings of the 23rd International Conference on
               Autonomous Agents and Multi-Agent Systems (AAMAS)},
  year      = {2024}
}

@article{chen2026multidiscounting,
  author  = {Chen, Pengbin and Liu, Qi and Li, Yanjie and Yan, Kejian and
             Ma, Shuangkang},
  title   = {Multi-Discounting Reinforcement Learning Based on Reward
             Decomposition},
  journal = {IEEE Transactions on Emerging Topics in Computing},
  volume  = {14},
  number  = {1},
  pages   = {1--14},
  year    = {2026}
}

@article{moghimi2026decoupling,
  author  = {Moghimi, Mehrdad and Coache, Anthony and Ku, Hyejin},
  title   = {Decoupling Time and Risk: Risk-Sensitive Reinforcement
             Learning with General Discounting},
  journal = {arXiv preprint arXiv:2602.04131},
  year    = {2026}
}

@article{ogawa2026adaptive,
  author  = {Ogawa, Kazuki and others},
  title   = {Adaptive Discount Factor for Accelerating Policy Learning
             Considering Long-Term Returns in Reinforcement Learning
             with Non-Stationary Environments},
  journal = {Journal of Computational Science},
  year    = {2026}
}

@inproceedings{ziebart2008maximum,
  author    = {Ziebart, Brian D. and Maas, Andrew and Bagnell, J. Andrew
               and Dey, Anind K.},
  title     = {Maximum Entropy Inverse Reinforcement Learning},
  booktitle = {Proceedings of the 23rd AAAI Conference on Artificial
               Intelligence},
  pages     = {1433--1438},
  year      = {2008}
}

@inproceedings{lakshminarayanan2017simple,
  author    = {Lakshminarayanan, Balaji and Pritzel, Alexander and
               Blundell, Charles},
  title     = {Simple and Scalable Predictive Uncertainty Estimation
               using Deep Ensembles},
  booktitle = {Advances in Neural Information Processing Systems},
  volume    = {30},
  year      = {2017}
}

@inproceedings{yoshida2013reinforcement,
  title={Reinforcement learning with state-dependent discount factor},
  author={Yoshida, Naoto and Uchibe, Eiji and Doya, Kenji},
  booktitle={2013 IEEE third joint international conference on development and learning and epigenetic robotics (ICDL)},
  pages={1--6},
  year={2013},
  organization={IEEE}
}

@article{tsitsiklis1994asynchronous,
  author  = {Tsitsiklis, John N.},
  title   = {Asynchronous Stochastic Approximation and Q-Learning},
  journal = {Machine Learning},
  volume  = {16},
  number  = {3},
  pages   = {185--202},
  year    = {1994}
}

\end{document}